\documentclass[,journal]{IEEEtran}
\usepackage{amsmath,amsfonts}
\usepackage{algorithmic}
\usepackage{array}
\usepackage[caption=false,font=normalsize,labelfont=sf,textfont=sf]{subfig}
\usepackage{textcomp}
\usepackage{stfloats}
\usepackage{url}
\usepackage{verbatim}
\usepackage{graphicx}
\usepackage{booktabs}

\usepackage{xcolor}
\usepackage{subfig}
\usepackage{pgfplots}
\pgfplotsset{compat=1.18} 
\usepackage{multirow}
\usepackage[table]{xcolor}
\usepackage{colortbl}
\def\BibTeX{{\rm B\kern-.05em{\sc i\kern-.025em b}\kern-.08em
    T\kern-.1667em\lower.7ex\hbox{E}\kern-.125emX}}
\usepackage{balance}
\usepackage{colortbl}
\definecolor{lightgray}{gray}{0.9}
\begin{document}
\title{ResGene-T: A Tensor-Based Residual Network Approach for Genomic Prediction}
\author{Kuldeep Pathak, Kapil Ahuja*, Eric de Sturler
\thanks{Kuldeep Pathak and Kapil Ahuja are with the Department of Computer Science and Engineering, Indian Institute of Technology Indore, India 453552 (e-mail: phd2301101009@iiti.ac.in; kahuja@iiti.ac.in). \\ *Corresponding author: Kapil Ahuja.}
\thanks{Eric de Sturler is with Department of Mathematics, Virginia Tech, Blacksburg, USA 24061 (e-mail: sturler@vt.edu).}
}

\maketitle

\begin{abstract}
In this work, we propose a new deep learning model for Genomic Prediction (GP), which involves correlating genotypic data with phenotypic. The genotypes are typically fed as a sequence of characters to the 1D-Convolution Neural Network layer of the underlying deep learning model. Inspired by earlier work that represented genotype as a 2D-image for genotype-phenotype classification, we extend this idea to GP, which is a regression task. We use a ResNet-18 as the underlying architecture, and term this model as ResGene-2D. 

Although the 2D-image representation captures biological interactions well, it requires all the layers of the model to do so. This limits training efficiency. Thus, as seen in the earlier work that proposed a 2D-image representation, our ResGene-2D performs almost the same as other models (3\% improvement). To overcome this, we propose a novel idea of converting the 2D-image into a 3D/ tensor and feed this to the ResNet-18 architecture, and term this model as ResGene-T.

We evaluate our proposed models on three crop species having ten phenotypic traits and compare it with seven most popular models (two statistical, two machine learning, and three deep learning). ResGene-T performs the best among all these seven methods (gains from $14.51$\% to $41.51$\%).
\end{abstract}



\section{Introduction}\label{sec:intro}
In traditional breeding, scientists select superior plants by observing their phenotype, such as plant height, grain yield, or disease resistance, and then use these selected plants as parents for the next generation. However, this approach requires waiting for plants to fully mature before evaluation can occur, which makes the breeding process time-consuming. Moreover, environmental factors like soil quality, water availability, and temperature can influence how traits are expressed, making it difficult to determine whether a plant's performance is driven by its genetics or simply favorable growing conditions~\cite{lorenz2011genomic}. 

Genetic data for a particular variety of a plant consist of a sequence of nucleotides, (which are bases/characters A, C, G, and T). The positions in the genetic data where nucleotides differ among varieties are defined as single nucleotide polymorphisms (SNPs). The biological interactions between SNPs are of great importance and influence how the plant grows.
Genomic prediction (GP) is a modern breeding approach that uses genetic data (genotype) of varieties to predict their quantitative physical traits (phenotype). This accelerated breeding process is particularly crucial in agriculture. 

Traditionally, GP methods can be broadly categorized into three groups: statistical methods, machine learning (ML) based methods, and deep learning (DL) based methods. Statistical approaches such as genomic best linear unbiased prediction (GBLUP), ridge regression best linear unbiased prediction (rrBLUP), BayesA, and BayesB have been used in the past works~\cite{meher2022performance,islam2023deepcgp,wu2023transformer}. ML based methods, notably Random Forest (RF)~\cite{ghosh2021enriched}, Support Vector Regression (SVR)~\cite{mao2025mdnn}, and Gradient Boosting Machines (GBM)~\cite{yan2021lightgbm}, have also been explored as promising alternatives. Similarly, DL based methods including DeepGS~\cite{ma2018deep}, DLGWAS~\cite{liu2019phenotype}, DNNGP~\cite{wang2023dnngp}, and GPFormer~\cite{lu2025soybean} have shown competitive results in recent studies.

All three categories of methods are competitive with each other, but none consistently turns out to be the best across all the datasets. Further, the performance of these methods, which is typically measured by Pearson Correlation Coefficient ($\mathcal{PCC}$), varies greatly and still leaves room for improvement. Thus, in this work, we propose a new DL based method that outperforms these methods on a variety of datasets. 

The goal of statistical, ML, and DL methods is to capture the underlying biological interactions among SNPs present in the genotype. DL based methods typically work by passing the genotype as a sequence of characters into the first layer of the model, which is usually a one dimensional convolutional neural network (1D-CNN). Since the genotype is fed one character at a time, these biological interactions among SNPs are difficult to capture in this setting. 

Hence, exploiting this idea, Muneeb et al.~\cite{muneeb2022can} for a genotype-phenotype case classification problem, proposed an approach of transforming genotype sequences into a 2D-image and feeding it to a 2D-CNN. This approach makes it easier to capture biological interactions among SNPs. We extend the idea of Muneeb et al.~\cite{muneeb2022can} from genotype-phenotype case classification problem to GP, which is a regression problem. There could be many underlying networks that can be used, but based on the need to avoid the vanishing gradient problem and maintain simplicity, we use ResNet-18. We term our new model as ResGene-2D. 

The results of Muneeb et al.~\cite{muneeb2022can} showed more stable performance across a variety of tests, however, the overall performance was not improved over the sequence of characters case ($3\%$ reduction). We see a similar behaviour. ResGene-2D gives performance comparable to other DL methods (3\% improvement). The reason for this behaviour is that although biological interactions are captured better when we change the input from a sequence of characters to a 2D-image, it still takes all the layers of the underlying model to learn these interactions (since the whole image is read in the last layer). Thus, the model is not trained effectively. To overcome this problem, we propose a novel idea of transforming the 2D-image into a 3D/tensor representation.

As before, we feed our 3D/ tensor image to a 2D-CNN. The multiple channels in the 2D-CNN layer capture the depth of our image. This process ensures that the whole genotype is read by the model in the initial layers itself, giving the model enough opportunity to learn the underlying biological relationships early-on/ well. This setup leads to substantial gains in performance, which we discuss below. The underlying model we use is again ResNet-18, and we term our model as ResGene-T. The results discussed below demonstrate that our model performs very well.

To comprehensively evaluate ResGene-2D and ResGene-T, we conduct extensive experiments with the following design:
\begin{itemize}
\item We evaluate our proposed model on three major crop species (soybean, rice, and sorghum), having ten phenotypic traits.
\item We compare ResGene-2D and ResGene-T against seven state-of-the-art methods, including two statistical approaches (rrBLUP and BayesB), two machine learning algorithms (SVR and XGBoost), and three recent deep learning models (DLGWAS, DNNGP, and GPFormer).
\item Recognizing that hyperparameter settings play a crucial role in deep learning performance, {\it we conduct extensive hyperparameter tuning} for all deep learning baselines to ensure a fair comparison (eight combinations).
\end{itemize}

As we have mentioned above, ResGene-T outperforms ResGene-2D by $11.85\%$. Further, ResGene-T outperforms the other seven existing methods, with improvements ranging from $14.51\%$ to $41.51\%$. We also use another measure to evaluate how well our method performs. The main advantage of our method is that it performs consistently well. If we look at ResGene-T, it ranks as the number one model in $7$ out of $10$ traits, and in the remaining $3$ traits, it achieves either the second or third position. Thus, our model, ResGene-T, has an average ranking of $1.4$. In comparison, the second and third best models, which are ResGene-2D and SVR, have average ranks of $3.5$ and $3.9$, respectively.

The rest of this manuscript consists of four more sections. In Section \ref{sec:ls}, we review the existing literature. The design of our ResGene-2D and ResGene-T models are described in Section~\ref{sec:alg}. In Section~\ref{sec:result}, we discuss the experimental results. Finally, the conclusion and the future work are given in Section~\ref{sec:con}.

\section{Related Works}\label{sec:ls}
In this section, we review key prior studies related to genomic prediction (GP) and outline the novelty of our work. Section~\ref{sec:sm} reviews existing studies on statistical methods. Prior work related to machine learning methods is summarized in Section~\ref{sec:ml}. Section~\ref{sec:dl} describes existing studies that have used deep learning models.

\subsection{Statistical Methods}\label{sec:sm}
Table~\ref{tab:ls_sm} provides a summary of existing work on statistical methods for GP. These studies evaluated different methods across various crops and traits.

Wang et al.~\cite{wang2018expanding} in 2018 introduced two new methods for GP and evaluated these methods on Arabidopsis, maize, mice, and rice datasets containing a total of one hundred fifty seven traits. The methods are super BLUP (SBLUP) and compressed BLUP (CBLUP). The authors compared their methods with GBLUP and Bayesian LASSO. Their results showed that SBLUP achieved superior performance for traits controlled by fewer genes, while CBLUP outperformed GBLUP in most cases.

Haile et al.~\cite{haile2021genomic} in 2021 applied a broad set of statistical methods for GP on a wheat dataset containing six traits. They compared rrBLUP, GBLUP, BayesB, Bayesian LASSO (BL), Bayesian reproducing kernel Hilbert spaces (RKHS), and a combined GS + de novo GWAS model. Their findings indicated that BayesB outperformed the other methods across most traits.

Meher et al.~\cite{meher2022performance} in 2022 applied a set of statistical methods for GP on wheat, maize, barley, and simulated data, consisting of nine traits in total. They compared GBLUP, compressed BLUP (CBLUP), super BLUP (SBLUP), BayesA, BayesB, BayesC, Bayes LASSO (BLASSO), and Bayes Ridge Regression (BRR). The results showed that GBLUP and CBLUP achieved the highest performance across most traits.

\begin{table}[!ht]
\centering
\caption{Summary of previous studies based on Statistical Methods}
\label{tab:ls_sm}
\begin{tabular}{|p{1.7cm}|p{2cm}|p{1cm}|p{2.5cm}|}
\hline
\textbf{Studies} & \textbf{Dataset} & \textbf{Traits} & \textbf{Methods} \\
\hline

Wang et al.~\cite{wang2018expanding} (2018) & Arabidopsis, Maize, Mice, Rice, Simulated data & One hundred fifty seven &
SBLUP, CBLUP, GBLUP, Bayesian LASSO \\
\hline

Haile et al.~\cite{haile2021genomic} (2021) & Wheat & Six &
rrBLUP, GBLUP, BayesB, BL, RKHS, GS + de novo GWAS \\
\hline

Meher et al.~\cite{meher2022performance} (2022) & Wheat, Maize, Barley, Simulated data & Nine &
GBLUP, CBLUP, SBLUP, BayesA, BayesB, BayesC, BLASSO, BRR \\
\hline

\end{tabular}

\end{table}

\subsection{Machine Learning Methods}\label{sec:ml}
Table~\ref{tab:ls_ml} summarizes previous studies that applied machine learning techniques for GP, which form an alternative to traditional statistical methods.

Grinberg et al.~\cite{grinberg2020evaluation} in 2020 conducted an evaluation of machine learning algorithms for GP and compared them with statistical methods. They used three datasets spanning yeast, rice, and wheat, comprising a total of sixty-two traits. The authors evaluated machine learning techniques, including Elastic Net, Ridge Regression, LASSO Regression, Random Forest (RF), Gradient Boosting Machine (GBM), and Support Vector Machine (SVM) with statistical baselines, GBLUP and Bloom. Their findings showed that machine learning methods outperformed statistical methods across the majority of datasets.

Yan et al.~\cite{yan2021lightgbm} in 2021 examined machine learning algorithms and compared them with statistical methods for GP using a maize dataset that included three traits. They evaluated the effectiveness of Light Gradient Boosting Machine (LightGBM) and compared it with rrBLUP, XGBoost, CatBoost, and GBM. Their results showed that LightGBM consistently delivered the best balance between computational efficiency and prediction performance.

Wang et al.~\cite{wang2022using} in 2022 investigated the effectiveness of a set of machine learning algorithms with statistical methods for GP, on a pig dataset containing two traits. They evaluated Support Vector Regression (SVR), Kernel Ridge Regression (KRR), Random Forest (RF), and AdaBoost.R2, and compared these methods against genomic BLUP (GBLUP), single-step GBLUP (ssGBLUP), and BayesHE. The results demonstrated that these machine learning methods outperformed GBLUP, ssGBLUP, and BayesHE.

\begin{table}[!ht]
\centering
\caption{Summary of previous studies based on Machine Learning Methods}
\label{tab:ls_ml}
\begin{tabular}{|p{1.7cm}|p{1.5cm}|c|p{2.5cm}|}
\hline
\textbf{Studies} & \textbf{Dataset} & \textbf{Traits} & \textbf{Methods} \\
\hline

Grinberg et al.~\cite{grinberg2020evaluation} (2020) & Yeast, Rice, Wheat & Sixty two &
Elastic net, Ridge Regression, LASSO Regression, RF, GBM, SVM, GBLUP, Bloom \\
\hline
Yan et al.~\cite{yan2021lightgbm} (2021) & Maize & Three & LightGBM, XGBoost, CatBoost, rrBLUP\\
\hline
Wang et al.~\cite{wang2022using} (2022) & Pigs & Two & SVR, KRR, RF, AdaBoost.R2, GBLUP, ssGBLUP, BayesHE  \\
\hline
\end{tabular}

\end{table}

\subsection{Deep Learning Methods}\label{sec:dl}

Table~\ref{tab:ls_dl} summarizes existing work on deep learning methods for GP. With the advancement of deep learning, which have shown improvements over machine learning methods, these approaches have been increasingly applied to this domain as alternatives to statistical and machine learning methods.

Liu et al.~\cite{liu2019phenotype} in 2019 introduced a model named DLGWAS for GP. They evaluated their model using soybean and simulated datasets containing five phenotypic traits and compared its performance against BayesA, Bayesian LASSO, Bayesian Ridge Regression (BRR), rrBLUP, DeepGS, and SingleCNN. The results demonstrated that DLGWAS outperformed most statistical methods across the examined traits.

Sehrawat et al.~\cite{sehrawat2023predicting} in 2023 proposed a deep convolutional neural network model named NovGMDeep. They evaluated their model using datasets of Arabidopsis thaliana and Oryza sativa, which included eight phenotypic traits. The model was compared against GBLUP, rrBLUP, G2PDeep, and DeepGS. Their results showed that NovGMDeep achieved higher prediction performance than both deep learning and statistical models.

Wang et al.~\cite{wang2023dnngp} in 2023 introduced a deep learning model called DNNGP. They evaluated their model using datasets of wheat, maize, and tomato, which together included nineteen phenotypic traits. The model was compared with GBLUP, SVR, LightGBM, DLGWAS, and DeepGS. Their findings indicated that DNNGP consistently surpassed these methods across multiple traits.

Lu et al.~\cite{lu2025soybean} in 2024 proposed GPformer, a Transformer-based model for GP. They evaluated their model on a soybean dataset containing three traits and compared it against Support Vector Regression (SVR), Random Forest (RF), XGBoost, and Multilayer Perceptron (MLP). The results revealed that GPformer achieved comparable prediction performance across all traits.

As discussed in the Introduction, the performance of all these methods could be further improved. To achieve this goal, we propose a novel ResNet-based technique referred to as ResGene.

\begin{table}[!ht]
\centering
\caption{Summary of previous studies based on Deep Learning Methods}
\label{tab:ls_dl}
\begin{tabular}{|p{1.7cm}|p{1.5cm}|c|p{2.5cm}|}
\hline
\textbf{Studies} & \textbf{Dataset} & \textbf{Traits} & \textbf{Methods} \\
\hline

Liu et al.~\cite{liu2019phenotype} (2019) & Soybean, Simulated data & Five &
DLGWAS, DeepGS, Single CNN, rrBLUP, BRR, BayesA, Bayessian LASSO \\
\hline

Sehrawat et al.~\cite{sehrawat2023predicting} (2023) & Arabidopsis Thaliana, Oryza Sativa & Eight &
NovGMDeep, G2Pdeep, DeepGS, GBLUP, rrBLUP  \\
\hline
Wang et al.~\cite{wang2023dnngp} (2023) & Wheat,
Maize, Tomato & Nineteen & DNNGP, DLGWAS, DeepGS, SVR, LightGBM, GBLUP\\
\hline
Lu et al.~\cite{lu2025soybean} (2024) & Soybean & Three & GPFormer, MLP, SVR, RF, XGBoost\\
\hline
\end{tabular}

\end{table}

\section{Our proposed ResGene model}\label{sec:alg}
This section is divided into three subsections. Section~\ref{sec:structdata} discusses structure of the data. The approach of transforming genotypic data into a 2D image and ResGene-2D are described in Section~\ref{sec:res2d}. Finally, Section~\ref{sec:res-t} provides details of our proposed ResGene-T model.

\subsection{Structure of the Data}\label{sec:structdata}

\begin{table}[ht]
\centering

\begin{tabular}{c|cccc}
\textbf{Varieties} & $\boldsymbol{\mathcal{M}_1}$ & $\boldsymbol{\mathcal{M}_2}$ & $\cdots$ & $\boldsymbol{\mathcal{M}_d}$ \\
\specialrule{.8pt}{0pt}{0pt}
$\mathcal{V}_1$ & A & K & $\cdots$ & T \\
$\mathcal{V}_2$ & A & K & $\cdots$ & C \\
$\vdots$ & $\vdots$ & $\vdots$ & $\vdots$ & $\vdots$ \\
$\mathcal{V}_n$ & R & G & $\cdots$ & T \\

\end{tabular}
\hspace{2pt}\vrule \vrule\hspace{2pt}
\begin{tabular}{c|c}
\textbf{PH} & \textbf{NN} \\ \specialrule{.8pt}{0pt}{0pt}
48.83 & 6.86 \\
54.04 & 10.65 \\
$\vdots$ & $\vdots$ \\
42.93 & 3.425 \\

\end{tabular}
\caption{Structure of the data}
\label{tab:data_summ}
\end{table}

We begin by describing the structure and organization of the datasets used in this study. Table~\ref{tab:data_summ} summarizes the key characteristics of dataset. The left part of this table denotes genotypic data, where rows $(\mathcal{V}_1 ... \mathcal{V}_n)$ correspond to the total of $n$ varieties (samples). The columns $(\mathcal{M}_1, \ldots, \mathcal{M}_d)$ represent $d$ Single Nucleotide Polymorphisms (SNPs), which serve as features for machine learning models. Each SNP column contains nucleotide values including $A$, $T$, $G$, $C$, $R$, $Y$, $S$, $W$, $K$, $M$, and missing values denoted by $N$. The right part of the table contains the corresponding phenotypic data for each variety. Here each column represents a specific trait, such as PH (Plant Height) and NN (Number of Nodes).

Next, we encode the genotypic data into a numerical format, as this is required for computational purposes. The encoding scheme is as follows: nucleotides $A$ and $T$ are encoded as $0$, while $G$ and $C$ are encoded as $2$. Missing values $N$ are encoded as $-1$, and all other values as $1$. We use this encoding scheme as it has been shown to be effective for genomic data~\cite{gaptiv3}.
\subsection{ResGene-2D}\label{sec:res2d}
\begin{figure*}[ht]
    \centering
    \includegraphics[width=0.9\linewidth]{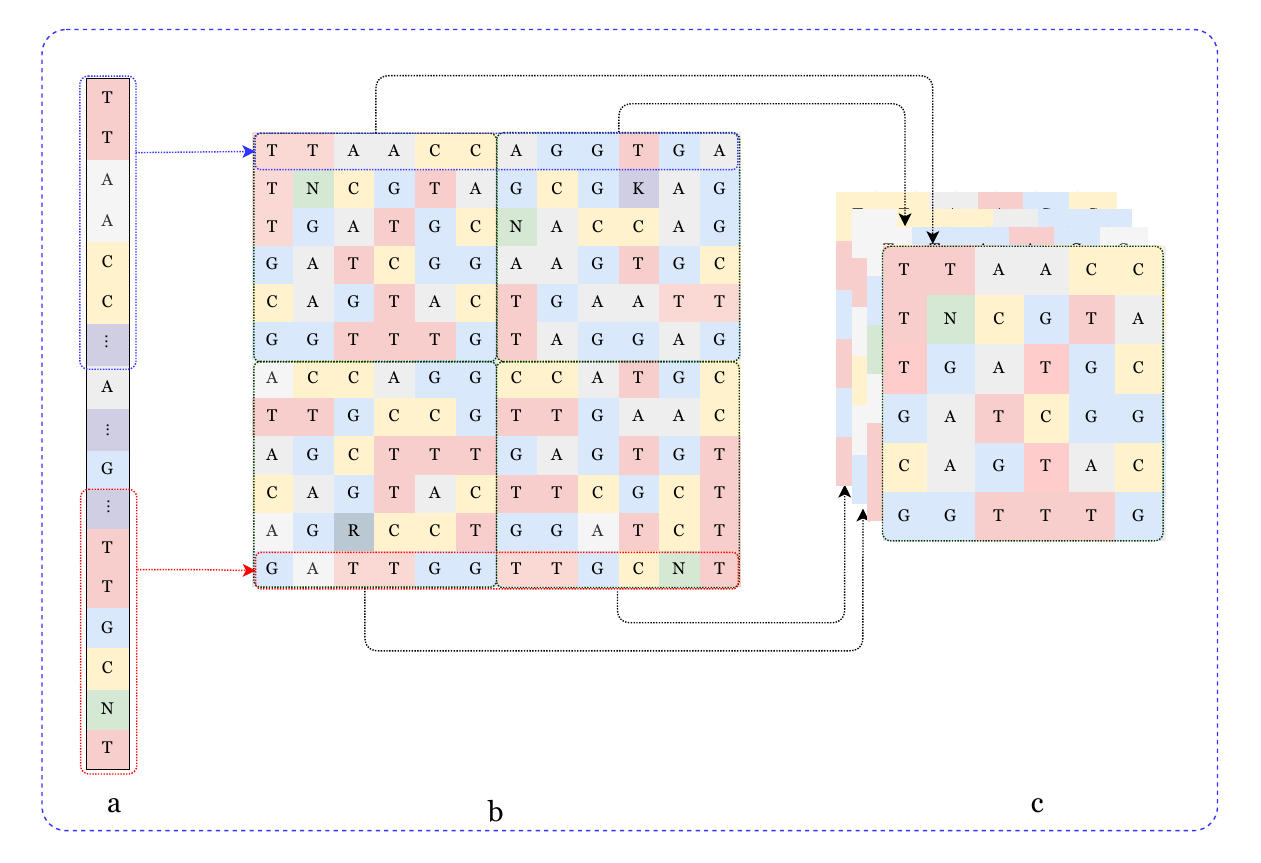}
    
    \caption{Pipeline for genotype to tensor image transformation. (a) Original genotype represented as a sequence of characters. (b) Conversion of the genotype sequence into a 2D-image. (c) Transformation of the 2D-image into a tensor representation.}

    \label{fig:snptimg}
\end{figure*}

As evident from the literature, deep learning models have been used in this domain. Two categories of models are commonly used, namely CNN-based methods and transformer-based methods. However, in all these approaches, genotypic data is fed into the model as a sequence of characters (Fig.~\ref{fig:snptimg}a), which is then passed to a 1D-CNN and correlated with the corresponding phenotypic value.

The goal of statistical, machine learning, and deep learning models is to capture certain biological interactions among SNPs. These interactions can be broadly categorized as linear (additive) and non-linear (epistatic). 

\begin{figure*}
    \centering
    \includegraphics[width=0.9\linewidth]{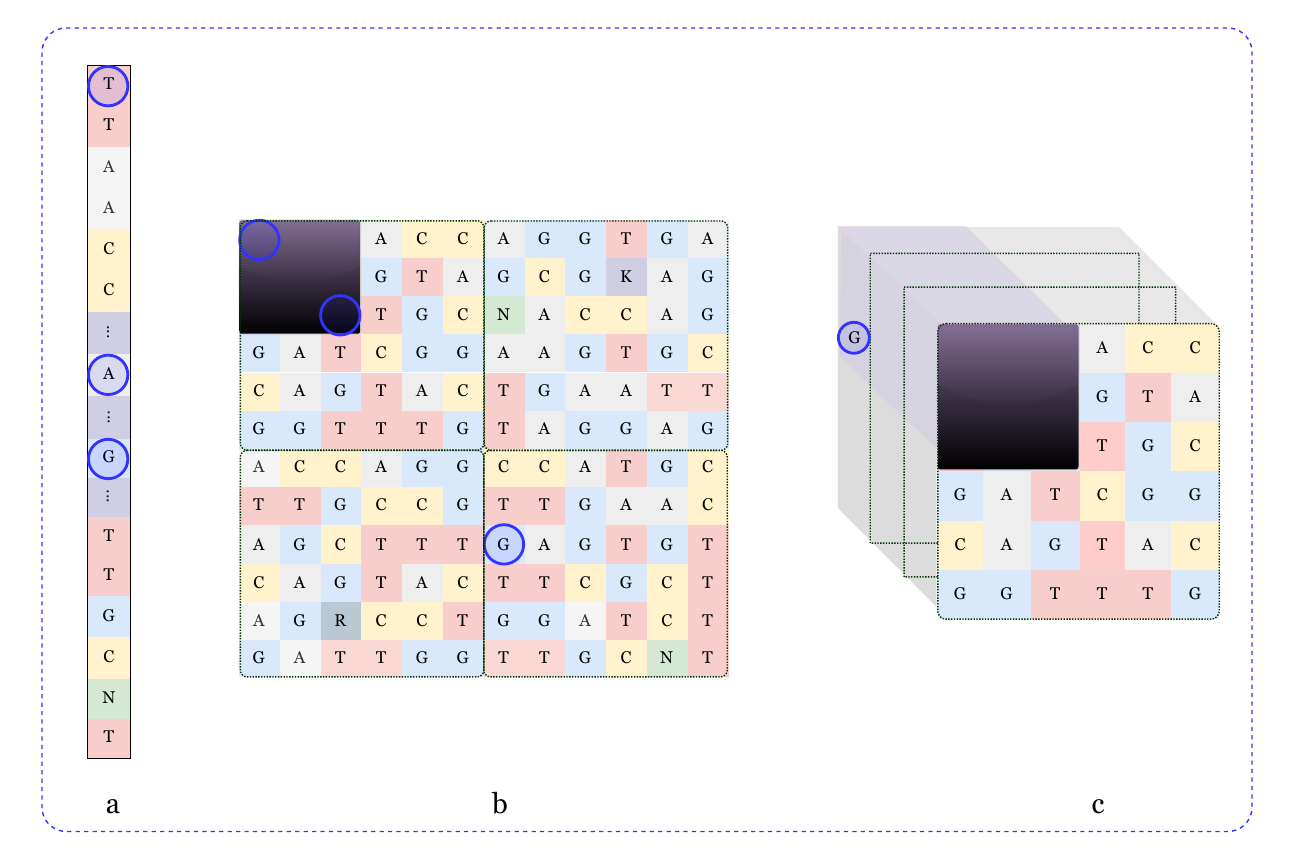}

    \caption{Biological interaction among different SNPs.}
    \label{fig:eps}
\end{figure*}

Although the approach of using a 1D-CNN in a deep learning model is intuitive, it has inherent drawbacks that it does not capture the above discussed biological interactions well. For example, see Fig.~\ref{fig:eps}a, where nucleotide $T$ and $A$ (circled in blue), which may have biological interaction, are far apart. Hence, to overcome this problem, it has been proposed in the literature to convert the genotype into a 2D-image, which may help to address this issue. The conversion of genotype into a 2D-image is shown in Fig.~\ref{fig:snptimg}b. When this 2D-image is fed to a 2D-CNN, the model reads the image using kernels in patches. Typically, for example, in the deep learning model which we use below, this kernel is of size $3\times3$. If we look at Fig.~\ref{fig:eps}b, the T and A nucleotides now fall within one kernel read, and hence their biological interaction can be captured more easily.

This idea was proposed by Muneeb et al.~\cite{muneeb2022can}. In broad terms, they argued that moving from a 1D-CNN to a 2D-CNN is favored due to improved stability, which may be linked to the better capture of these biological interactions. They applied this approach to a genotype-phenotype case classification problem. In this work, we translate this approach to GP, which is a regression problem.

Several CNN architectures have been proposed for deep learning tasks~\cite{feng2023survey}. These architectures can be broadly categorized into two groups based on their ability to handle the vanishing gradient problem. The first group includes early architectures such as AlexNet, VGG, and InceptionNet, which struggle with vanishing gradients in deeper networks. The second group comprises more recent architectures like Residual Network (ResNet), DenseNet, and EfficientNet, which effectively address this limitation through novel design principles.

Among these modern architectures, ResNet is simpler and more balanced, hence we use it as backbone for our model. Since DenseNet offers denser connectivity patterns that could be explored in future work, its increased complexity is not necessary for our current objectives. Similarly, EfficientNet optimizes for computational efficiency, which is not a primary constraint in our study. 

\begin{figure*}
    \centering
    \includegraphics[width=\linewidth]{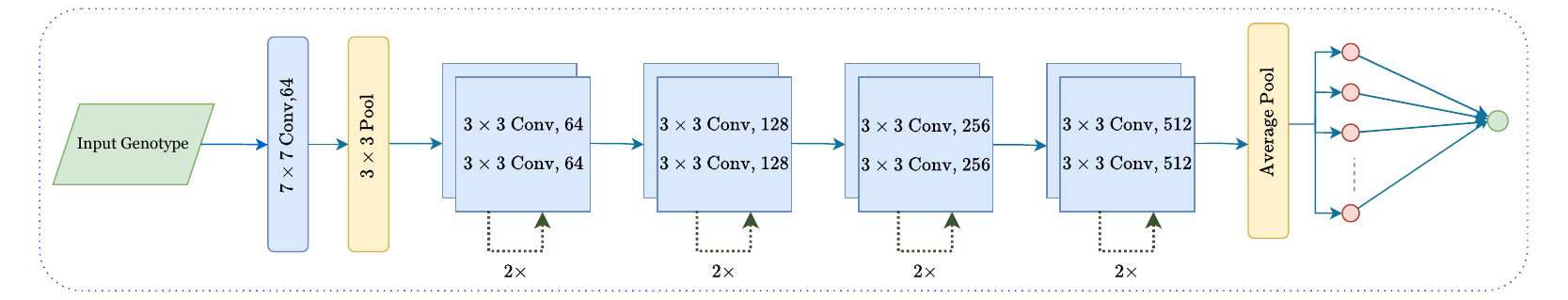}
    \caption{ResNet-18 architecture.}
    \label{fig:main_arch}
\end{figure*}

ResNet architectures are typically pre-trained on ImageNet~\cite{pandey2023resnet} for computer vision tasks. Since we are using genomic data, we train ResNet from scratch. Various ResNet variants exist, including ResNet-18, ResNet-34, ResNet-50, ResNet-101, and ResNet-152. We select ResNet-18 as our base architecture due to the small size of our genomic dataset. Following the feature extraction, we modify final fully connected layer of the model to adapt it for a regression task. Specifically, we replace it with a linear layer that maps the extracted features to a single output value. Since our model is based on the ResNet-18 framework and designed for genomic applications, we refer to it as \textbf{ResGene-2D}. The complete architecture of this model is illustrated in Fig.~\ref{fig:main_arch}.

\subsection{ResGene-T}\label{sec:res-t}

The performance of 2D-CNN as reported by Muneeb et al.~\cite{muneeb2022can} turns out to be slightly worse, showing about $3\%$ deterioration compared to that of 1D-CNN. Our results, which we discuss in the next section, are consistent with this observation, where ResGen-2D performs similarly to the existing deep learning models cited above (a 3\% improvement).

The main reason for this is that although 2D-CNN is able to capture biological interactions better, it requires many CNN layers to do so (since the whole image is read in the last layer). Thus, the model is not trained well. To overcome this problem, we propose a novel approach of transforming the 2D-image into a 3D/tensor representation, as shown in Fig.~\ref{fig:snptimg}c. Here, the first layer of the model is still a 2D-CNN with a kernel size of $3\times3$, which reads the 2D-images, and $4$ channels, which read each of the 2D-images in-parallel.

Now, let us observe the biological interactions again while referring to Fig.~\ref{fig:eps}. Earlier in 1D case (Fig.~\ref{fig:eps}a) SNPs $T$ and $A$, were far apart, which had come in the same kernel read in 2D case (Fig.~\ref{fig:eps}b). This behavior gets replicated when we go to tensor case (Fig.~\ref{fig:eps}c), where SNPs $T$ and $A$ fall in the same kernel read.

Further, if we now observe SNPs $T$ and $G$, they were very far apart in the 1D case (Fig.~\ref{fig:eps}a) and came only slightly closer in the 2D case (Fig.~\ref{fig:eps}b), however, in the tensor representation, these SNPs fall within the same kernel read (Fig.~\ref{fig:eps}c). This means that the biological interactions of a much larger number of SNPs are brought together much earlier in the network, and hence, the model should be able to capture the correlation between genotypic and phenotypic data more effectively. Next, we describe the mathematical relationship that quantifies how much earlier the entire genotype is read when moving from a 2D image representation to a tensor representation.

\begin{table}[!ht]
 \renewcommand{\arraystretch}{1.8} 
\centering 
\caption{Comparison of CNN layer requirements for 2D image and tensor representations}
\label{tab:eq_l}
\begin{tabular}{|c c||c|}
\hline
 & \textbf{2D-Image Representation} & \textbf{Tensor Representation} \\
\hline
1. & $\mathcal{S}_{2D} \times \mathcal{S}_{2D} = d$ 
& $\mathcal{S_T} \times \mathcal{S_T} \times \mathcal{C} = d$ \\[6pt]
\hline
2. & $\mathcal{S}_{2D} = \sqrt{d}$ 
& $\mathcal{S_T} = \sqrt{\frac{d}{\mathcal{C}}}$ \\[6pt]
\hline
3. & $L_{2D} \approx \frac{\mathcal{S}_{2D}}{k} = \frac{\sqrt{d}}{k}$ 
& $L_{T} \approx \frac{\mathcal{S_T}}{k} = \frac{\sqrt{d}}{k\sqrt{\mathcal{C}}}$ \\[8pt]
\hline
4. & \multicolumn{2}{c|}{\textbf{Ratio:} $\dfrac{L_{2D}}{L_{T}} = \sqrt{\mathcal{C}}$} \\[6pt]
\hline
\end{tabular}

\end{table}

Table~\ref{tab:eq_l} provides a mathematical comparison of the convolutional layers required to achieve full genotype coverage in 2D-image versus tensor representations. We now explain this table row-wise. In row 1, the 2D-image of size $\mathcal{S}_{2D} \times \mathcal{S}_{2D}$ has $\mathcal{S}_{2D}$ as the dimension of square image and $d$ as the total number of SNPs. In the same row, the tensor image of size $\mathcal{S_T} \times \mathcal{S_T} \times \mathcal{C}$ has $\mathcal{S}_{T}$ as the dimension of square image, $\mathcal{C}$ as the number of channels, and $d$ as earlier, as the total number of SNPs.  

Row 2 gives the size of the square image in both cases when $d$ and $\mathcal{C}$ are fixed. Row 3 discusses the number of CNN layers required to cover the entire image, where $k$ is the kernel size. Finally, row 4 gives the ratio of number of layers required to cover the full image in 2D versus tensor case. As evident, the number of layers required in the 2D case is $\sqrt{\mathcal{C}}$ times more. Hence, the entire genotype is read earlier in the tensor case than in the 2D case.

\section{Result}\label{sec:result}
This section is organized into five subsections. The datasets used in this study are described in Section~\ref{subsec:dataset}. Section~\ref{subsec:base_model} discusses the baseline model, which we compare against. Tunning and experimental setup are given in Section~\ref{subsec:tun_set}. In Section~\ref{subsec:ev_me} we describe the evalution metric used in this study. Finally the genomic prediction performance is given in the Section~\ref{subsec:res}.

\subsection{Datasets}\label{subsec:dataset}
As discussed earlier, our model is applicable to data from any plant species. In this study, we use three distinct datasets from major crop species to ensure a comprehensive analysis. The first dataset we use is Soybean consisting of $269$ varieties~\cite{icarNSRI}. This includes $66,589$ Single Nucleotide Polymorphisms (SNPs) and corresponding phenotypic traits are Plant Height (PH), Number of Nodes (NN), Grain Yield (GY), and Canopy Temperature (CT).

The second dataset we use is Rice consisting of $327$ varieties~\cite{azodi2019benchmarking}. This includes $57,542$ SNPs and corresponding traits are Plant Height (PH), Flowering Time (FT), and Grain Yield (GY). The third dataset we use is Sorghum consisting of $451$ varieties~\cite{azodi2019benchmarking}. This includes $56,299$ SNPs and corresponding phenotypic traits are Plant Height (PH), Moisture (MO), and  Grain Yield (GY).

\subsection{Baseline Models}\label{subsec:base_model}
\begin{figure*}[ht]
    \centering
    \begin{tikzpicture}
 \pgfplotscreateplotcyclelist{mycolorlist}{
    {blue!60!white,    fill=blue!60!white},
    {red!60!white,     fill=red!60!white},
    {teal!60!white,    fill=teal!60!white},
    {orange!60!white,  fill=orange!60!white},
    {violet!60!white,  fill=violet!60!white},
    {cyan!40!white,    fill=cyan!40!white},
    {olive!60!white,   fill=olive!60!white},
    {magenta!60!white, fill=magenta!60!white},
    {gray!60!white, fill=gray!60!white}
}
    \begin{axis}[
        ybar,
        cycle list name=mycolorlist,
        bar width=8pt,
        width=18cm,
        height=8cm,
        ymin=0,
        ymax=0.65,
        ytick={0,0.1,0.2,0.3,0.4,0.5},
        ylabel={$\mathcal{PCC}$},
        xlabel={Traits},
        symbolic x coords={PH,NN,GY,CT},
        xtick=data,
        enlarge x limits=0.15,
        nodes near coords,
        legend style={
            at={(0.5,1.08)},
            anchor=south,
            legend columns=9,
            /tikz/every even column/.append style={column sep=3pt}
        },
        nodes near coords style={
            font=\footnotesize,
            text=black,
            rotate=90,
            anchor=west,
            /pgf/number format/.cd,
                fixed,
                precision=4
        }
    ]
    \addplot coordinates {(PH,0.331) (NN,0.3422) (GY,0.3208) (CT,0.079)};
    \addlegendentry{rrBLUP};

    \addplot coordinates {(PH,0.3236) (NN,0.3345) (GY,0.3151) (CT,0.0159)};
    \addlegendentry{BayesB};

    \addplot coordinates {(PH,0.4210) (NN,0.3760) (GY,0.2010) (CT,0.0970)};
    \addlegendentry{SVR};

    \addplot coordinates {(PH,0.4348) (NN,0.2294) (GY,0.0857) (CT,0.0439)};
    \addlegendentry{XGBoost};

    \addplot coordinates {(PH,0.32383) (NN,0.28409) (GY,0.19555) (CT,0.11725)};
    \addlegendentry{DLGWAS};

    \addplot coordinates {(PH,0.3634) (NN,0.3109) (GY,0.1568) (CT,0.0881)};
    \addlegendentry{DNNGP};
    \addplot coordinates {(PH,0.3773) (NN,0.3563) (GY,0.2232) (CT,0.1404)};
    \addlegendentry{GPFormer};
    \addplot coordinates {(PH,0.4465) (NN,0.3613) (GY,0.2910) (CT,0.1573)};
    \addlegendentry{ResGene-2D};
    \addplot coordinates {(PH,0.4675) (NN,0.3869) (GY,0.3011) (CT,0.1453)};
    \addlegendentry{ResGene-T};
    \end{axis}
    \end{tikzpicture}
    \caption{Model Performance on Soybean Dataset.}
    \label{fig:bar_icar}
\end{figure*}
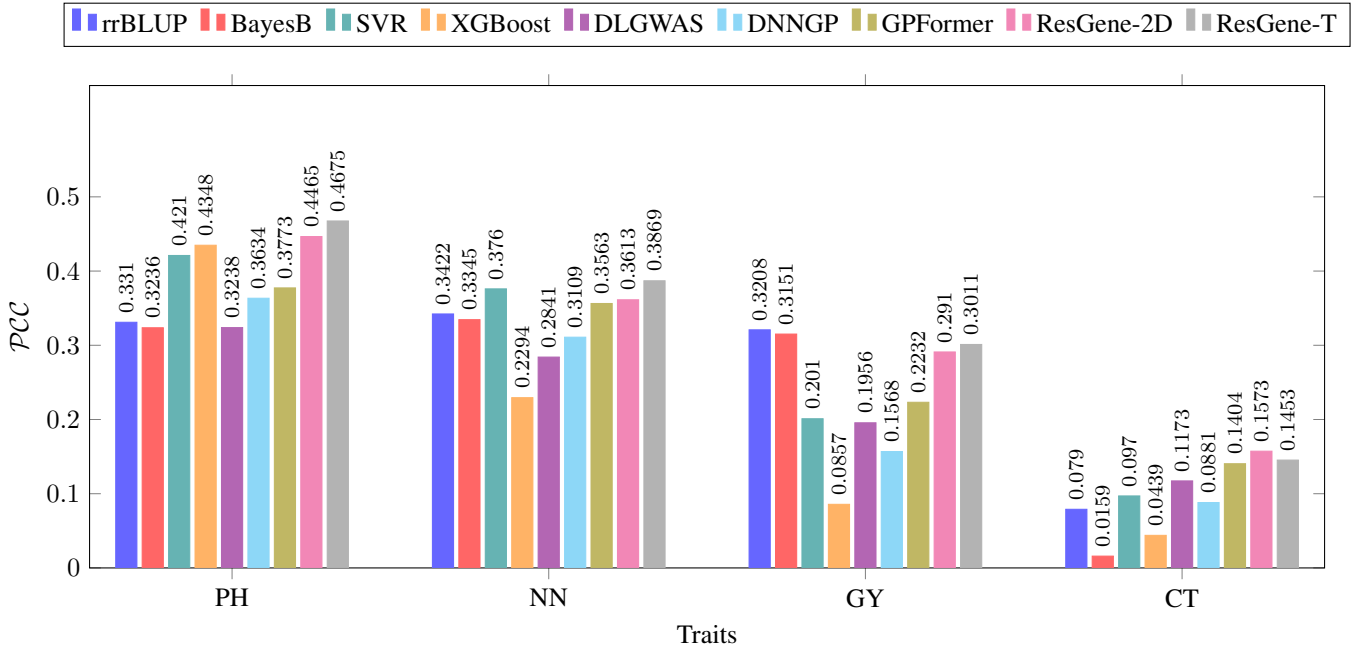

\begin{table*}[t]
\centering
\caption{Optimal Batch Size (BS), Learning Rate (LR), and Dropout (D) configurations for DLGWAS, DNNGP, and GPFormer}
\label{tab:h_opti}
\small
\setlength{\tabcolsep}{12pt}
\renewcommand{\arraystretch}{1.4}
\begin{tabular}{|l|l|c|c|c|}

\hline
\textbf{Dataset} & \textbf{Trait} & \textbf{DLGWAS} & \textbf{DNNGP} & \textbf{GPFormer}\\
\hline

\multirow{4}{*}{Soybean}
 & PH & BS:64, LR:0.001, D:0.3 & BS:32, LR:0.01, D:0.1 & BS:32, LR:0.001, D:0.3  \\
 & NN & BS:32, LR:0.001, D:0.3 & BS:32, LR:0.01, D:0.3 & BS:64, LR:0.001, D:0.3  \\
 & GY & BS:32, LR:0.001, D:0.3 & BS:32, LR:0.01, D:0.3 & BS:32, LR:0.001, D:0.3 \\
 & CT & BS:64, LR:0.001, D:0.3 & BS:32, LR:0.01, D:0.1 & BS:32, LR:0.001, D:0.3 \\
\hline

\multirow{3}{*}{Rice}
 & PH & BS:64, LR:0.001, D:0.3 & BS:64, LR:0.01, D:0.3 &BS:64, LR:0.001, D:0.3 \\
 & FT & BS:64, LR:0.001, D:0.1 & BS:64, LR:0.01, D:0.3 & BS:64, LR:0.001, D:0.1  \\
 & GY & BS:64, LR:0.001, D:0.1 & BS:64, LR:0.01, D:0.3 & BS:64, LR:0.001, D:0.1  \\
\hline

\multirow{3}{*}{Sorghum}
 & PH & BS:64, LR:0.001, D:0.1 & BS:32, LR:0.01, D:0.3 & BS:32, LR:0.01, D:0.1  \\
 & MO & BS:64, LR:0.001, D:0.3 & BS:32, LR:0.01, D:0.3 & BS:64, LR:0.001, D:0.3  \\
 & GY & BS:64, LR:0.001, D:0.3 & BS:32, LR:0.01, D:0.3 & BS:64, LR:0.001, D:0.3  \\
\hline

\end{tabular}

\end{table*}

\begin{table*}[t]
\centering
\caption{Optimal Batch Size (BS), Learning Rate (LR), Dropout (D), and Channel ($\mathcal{C}$) configurations for ResGene-2D, and ResGene-T}
\label{tab:h_opti_res}
\small
\setlength{\tabcolsep}{12pt}
\renewcommand{\arraystretch}{1.4}
\begin{tabular}{|l|l|c|c|}
\hline
\textbf{Dataset} & \textbf{Trait} & \textbf{ResGene-2D} & \textbf{ResGene-T}\\
\hline

\multirow{4}{*}{Soybean}
 & PH & BS:32, LR:0.001, D:0.3 & BS:64, LR:0.001, D:0.1, $\mathcal{C}$:20 \\
 & NN & BS:64, LR:0.001, D:0.3 & BS:64, LR:0.001, D:0.1, $\mathcal{C}$:50 \\
 & GY & BS:64, LR:0.001, D:0.1 & BS:64, LR:0.001, D:0.3, $\mathcal{C}$:50 \\
 & CT & BS:64, LR:0.001, D:0.3 & BS:64, LR:0.001, D:0.3, $\mathcal{C}$:20 \\
\hline

\multirow{3}{*}{Rice}
 & PH & BS:32, LR:0.001, D:0.3 & BS:64, LR:0.001, D:0.1, $\mathcal{C}$:50 \\
 & FT & BS:32, LR:0.001, D:0.3 & BS:32, LR:0.001, D:0.3, $\mathcal{C}$:50 \\
 & GY & BS:32, LR:0.001, D:0.3 & BS:32, LR:0.001, D:0.1, $\mathcal{C}$:50 \\
\hline

\multirow{3}{*}{Sorghum}
 & PH & BS:32, LR:0.001, D:0.3 & BS:32, LR:0.001, D:0.3, $\mathcal{C}$:50 \\
 & MO & BS:32, LR:0.001, D:0.3 & BS:32, LR:0.001, D:0.1, $\mathcal{C}$:50 \\
 & GY & BS:32, LR:0.001, D:0.3 & BS:32, LR:0.001, D:0.3, $\mathcal{C}$:50 \\
\hline

\end{tabular}
\end{table*}

For a more robust and comprehensive comparison, we evaluate ResGene-2D and ResGene-T against seven widely used baseline models spanning statistical, machine learning, and deep learning approaches. The statistical models rrBLUP and BayesB are standard benchmarks in genomic prediction, and have been extensively used in prior studies~\cite{wang2025wheatgp, islam2023deepcgp}. To represent traditional machine learning methods, we include SVR and XGBoost, both of which have demonstrated strong predictive performance in recent genomic prediction works~\cite{wu2023transformer, wang2025wheatgp}. Finally, to ensure a fair comparison within the deep learning domain, we incorporate DLGWAS and DNNGP, two state-of-the-art deep learning models and more recent GPFormer. These models are specifically designed for genomic prediction task~\cite{wang2023dnngp, mao2025mdnn, lu2025soybean}. This selection provides a diverse set of baselines for evaluating our proposed ResGene-2D and ResGene-T models.

\subsection{Tuning and Setup}\label{subsec:tun_set}
The statistical models rrBLUP and BayesB are implemented in R. For rrBLUP, we follow standard practice~\cite{wang2025wheatgp} and fit the model using the $mixed.solve$ function after removing SNPs with zero or undefined variance. The rrBLUP model does not require tuning parameters, as regularization is determined automatically through variance component estimation. For BayesB, we use $10{,}000$ iterations with a burn-in period of $2{,}000$ cycles~\cite{zhang2010best}.

For machine learning models SVR and XGBoost, we use the scikit-learn Python library for implementation~\cite{wang2025wheatgp}. For SVR, we use linear kernel, with settings including the regularization parameter $C = 1.0$, $\epsilon = 0.1$, and a tolerance of $10^{-3}$. For XGBoost, we use default parameter settings, including the tree-based booster, a maximum tree depth of $6$, and an $\ell_2$ regularization parameter $\lambda = 1$.

For the deep learning models DLGWAS, DNNGP, and GPFormer, we use their publicly available implementations for experimentation. ResGene-2D and ResGene-T are implemented using PyTorch library in Python. The Stochastic Gradient Descent (SGD) optimiser is used and the Mean Squared Error (MSE) use as the loss function. All five DL models are trained for 100 epochs. 

The hyperparameter tuning for deep learning models presents a more complex challenge. Identifying optimal configurations for these models is inherently difficult due to their sensitivity to parameter choices and the vast search space involved. Our literature review revealed that existing studies typically report results using a single hyperparameter configuration for each deep learning model. We consider this approach potentially unfair for comparative evaluation, as it may not reflect the true performance potential of each model. Therefore, to ensure a rigorous and equitable comparison, we establish a systematic hyperparameter tuning protocol. We apply this protocol to baseline models (DLGWAS, DNNGP, and GPFormer) and our proposed models (ResGene-2D and ResGene-T).

We tune these models by focusing on three critical hyperparameters: Batch Size (BS), Learning Rate (LR), and Dropout (D). For each hyperparameter, we consider two widely used values: BS $\in \{32, 64\}$, LR $\in \{0.01, 0.001\}$, and D $\in \{0.1, 0.3\}$, drawn from the genomic prediction literature~\cite{liu2019phenotype, wang2023dnngp, wu2023transformer}. This yields a total of eight possible configurations for DLGWAS, DNNGP, GPFormer, and ResGene-2D. However, for ResGene-T model, we additionally tune the number of channels $\mathcal{C} \in \{20, 50\}$, which is specific to the tensor representation. This results in sixteen configurations for ResGene-T. 

We conduct extensive experiments evaluating all configurations for each deep learning model across all traits within each dataset. For each model, we select the configuration that achieves the best predictive performance. The optimal hyperparameters for the baseline models (DLGWAS, DNNGP, and GPFormer) are reported in Table~\ref{tab:h_opti}, while those for our proposed models (ResGene-2D and ResGene-T) are reported in Table~\ref{tab:h_opti_res}.

All experiments are conducted on a single machine (Intel Xeon CPU @ 2.00GHz, 128GB RAM). The evaluation uses 10-fold cross-validation, partitioning the dataset into ten equal folds with each fold serving once as the test set.

\subsection{Evaluation Metric}\label{subsec:ev_me}
Several evaluation metrics are commonly used for regression tasks, including mean squared error (MSE), mean absolute error (MAE), the coefficient of determination (R), and Pearson’s correlation coefficient ($\mathcal{PCC}$). In GP studies, many prior works primarily report $\mathcal{PCC}$ as the performance metric~\cite{liu2019phenotype, wang2023dnngp,mao2025mdnn}. Following this established practice, we also adopt $\mathcal{PCC}$ to evaluate the prediction performance of different models. It is defined as
\begin{equation}
\mathcal{PCC} =
\frac{\sum(x-\bar{x})(y-\bar{y})}
     {\sqrt{\sum(x-\bar{x})^{2}}
      \sqrt{\sum(y-\bar{y})^{2}}},
      \label{eq:pearson}
\end{equation}
where $x$ and $y$ denote the observed and predicted values, respectively, while $\bar{x}$ and $\bar{y}$ are the means of the observed and predicted values. It ranges from $-1$ to $+1$. A higher $\mathcal{PCC}$ value indicates a stronger linear relationship between genotypic and phenotypic data.

While $\mathcal{PCC}$ quantifies prediction accuracy, it does not indicate whether observed differences among models are statistically meaningful. To address this, we use the Friedman test~\cite{demvsar2006statistical}, a non-parametric procedure for comparing multiple models across multiple traits. For each trait, models are ranked according to their $\mathcal{PCC}$ values i.e. for a model $j \in {1, \dots, M}$ and trait $i \in {1, \dots, N}$ the rank is denoted by $r_i^j$. After this the average rank for a $j^{th}$ model is calculated as follows: 
\begin{equation}
R_j = \frac{1}{N}\sum_{i=1}^{N} r_i^j.
\end{equation}

The Friedman test evaluates the null hypothesis that all models have equal average ranks using the statistic,
\begin{equation}
\chi_F^2 = \frac{12N}{M(M+1)}\left(\sum_{j=1}^{M} R_j^2 - \frac{M(M+1)^2}{4}\right).
      \label{eq:freidman}
\end{equation}

The $p$-value corresponding to $\chi_F^2$ is computed from the table. A $p$-value less than $0.05$ indicates that the performance rankings differ significantly beyond random variation~\cite{andrade2019p}.

\subsection{Performance}\label{subsec:res}

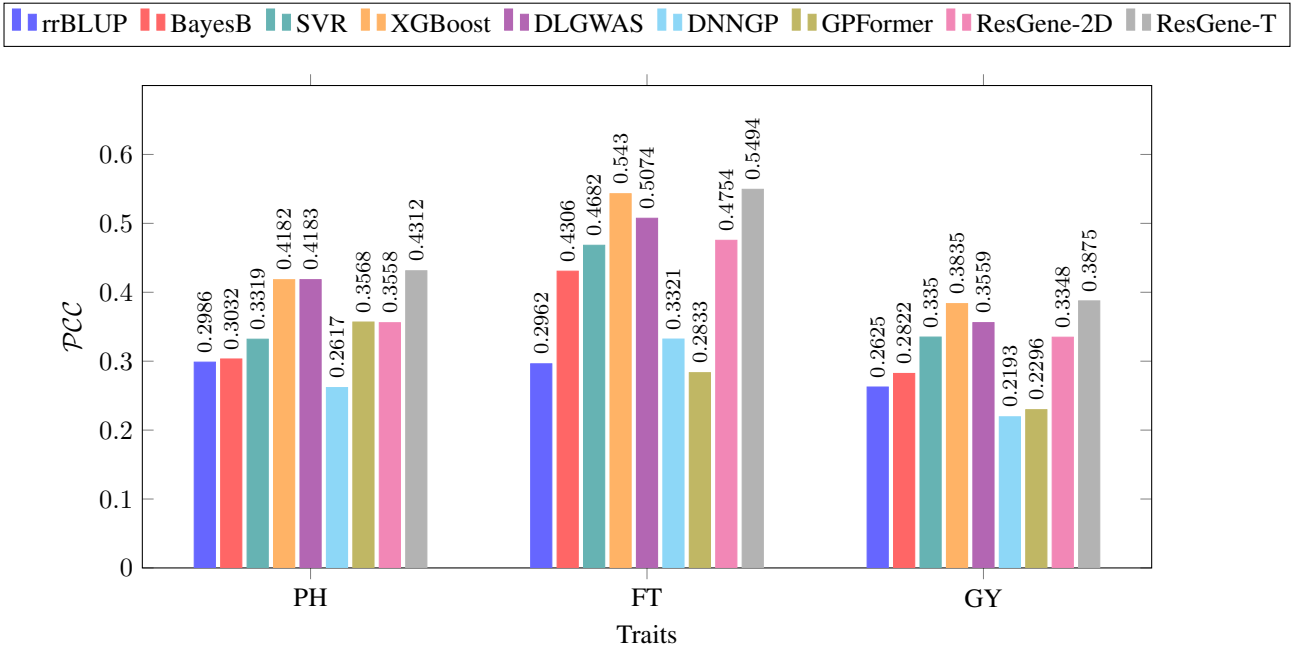
\begin{figure*}[ht]
    \centering
    \begin{tikzpicture}
 \pgfplotscreateplotcyclelist{mycolorlist}{
    {blue!60!white,    fill=blue!60!white},
    {red!60!white,     fill=red!60!white},
    {teal!60!white,    fill=teal!60!white},
    {orange!60!white,  fill=orange!60!white},
    {violet!60!white,  fill=violet!60!white},
    {cyan!40!white,    fill=cyan!40!white},
    {olive!60!white,   fill=olive!60!white},
    {magenta!60!white, fill=magenta!60!white},
    {gray!60!white, fill=gray!60!white}
}
    \begin{axis}[
        ybar,
        cycle list name=mycolorlist,
        bar width=8pt,
        width=15cm,
        height=8cm,
        ymin=0,
        ymax=0.70,
        ytick={0,0.1,0.2,0.3,0.4,0.5,0.6},
        ylabel={$\mathcal{PCC}$},
        xlabel={Traits},
        symbolic x coords={PH,FT,GY},
        xtick=data,
        enlarge x limits=0.25,
        nodes near coords,
        legend style={
            at={(0.5,1.08)},
            anchor=south,
            legend columns=9,
            /tikz/every even column/.append style={column sep=3pt}
        },
        nodes near coords style={
            font=\footnotesize,
            text=black,
            rotate=90,
            anchor=west,
            /pgf/number format/.cd,
                fixed,
                precision=4
        }
    ]
\addplot coordinates {(PH,0.2986) (FT,0.2962) (GY,0.2625)};
\addlegendentry{rrBLUP};

\addplot coordinates {(PH,0.3032) (FT,0.4306) (GY,0.2822)};
\addlegendentry{BayesB};

\addplot coordinates {(PH,0.3319) (FT,0.4682) (GY,0.3350)};
\addlegendentry{SVR};

\addplot coordinates {(PH,0.4182) (FT,0.5430) (GY,0.3835)};
\addlegendentry{XGBoost};

\addplot coordinates {(PH,0.4183) (FT,0.50739) (GY,0.35594)};
\addlegendentry{DLGWAS};

\addplot coordinates {(PH,0.2617) (FT,0.3321) (GY,0.2193)};
\addlegendentry{DNNGP};

\addplot coordinates {(PH,0.3568) (FT,0.2833) (GY,0.2296)};
\addlegendentry{GPFormer};

\addplot coordinates {(PH,0.3558) (FT,0.4754) (GY,0.3348)};
\addlegendentry{ResGene-2D};

\addplot coordinates {(PH,0.4312) (FT,0.5494) (GY,0.3875)};
\addlegendentry{ResGene-T};

    \end{axis}
    \end{tikzpicture}
    \caption{Model Performance on Rice Dataset.}
    \label{fig:bar_rice}
\end{figure*}

\begin{figure*}[ht]
    \centering
    \begin{tikzpicture}
 \pgfplotscreateplotcyclelist{mycolorlist}{
    {blue!60!white,    fill=blue!60!white},
    {red!60!white,     fill=red!60!white},
    {teal!60!white,    fill=teal!60!white},
    {orange!60!white,  fill=orange!60!white},
    {violet!60!white,  fill=violet!60!white},
    {cyan!40!white,    fill=cyan!40!white},
    {olive!60!white,   fill=olive!60!white},
    {magenta!60!white, fill=magenta!60!white},
    {gray!60!white, fill=gray!60!white}
}
    \begin{axis}[
        ybar,
        cycle list name=mycolorlist,
        bar width=8pt,
        width=15cm,
        height=8cm,
        ymin=0,
        ymax=0.80,
        ytick={0,0.1,0.2,0.3,0.4,0.5,0.6,0.7},
        ylabel={$\mathcal{PCC}$},
        xlabel={Traits},
        symbolic x coords={PH,MO,GY},
        xtick=data,
        enlarge x limits=0.25,
        nodes near coords,
        legend style={
            at={(0.5,1.08)},
            anchor=south,
            legend columns=9,
            /tikz/every even column/.append style={column sep=3pt}
        },
        nodes near coords style={
            font=\footnotesize,
            text=black,
            rotate=90,
            anchor=west,
            /pgf/number format/.cd,
                fixed,
                precision=4
        }
    ]
\addplot coordinates {(PH,0.5165) (MO,0.4922) (GY,0.2893)};
\addlegendentry{rrBLUP};

\addplot coordinates {(PH,0.5225) (MO,0.4938) (GY,0.2876)};
\addlegendentry{BayesB};

\addplot coordinates {(PH,0.5592) (MO,0.5477) (GY,0.4015)};
\addlegendentry{SVR};

\addplot coordinates {(PH,0.5242) (MO,0.4804) (GY,0.3597)};
\addlegendentry{XGBoost};

\addplot coordinates {(PH,0.6050) (MO,0.53964) (GY,0.3520)};
\addlegendentry{DLGWAS};

\addplot coordinates {(PH,0.5211) (MO,0.4365) (GY,0.3354)};
\addlegendentry{DNNGP};

\addplot coordinates {(PH,0.4197) (MO,0.4542) (GY,0.2761)};
\addlegendentry{GPFormer};

\addplot coordinates {(PH,0.5359) (MO,0.4986) (GY,0.3710)};
\addlegendentry{ResGene-2D};

\addplot coordinates {(PH,0.6031) (MO,0.5840) (GY,0.4281)};
\addlegendentry{ResGene-T};

    \end{axis}
    \end{tikzpicture}
    \caption{Model Performance on Sorghum Dataset.}
    \label{fig:bar_sor}
\end{figure*}
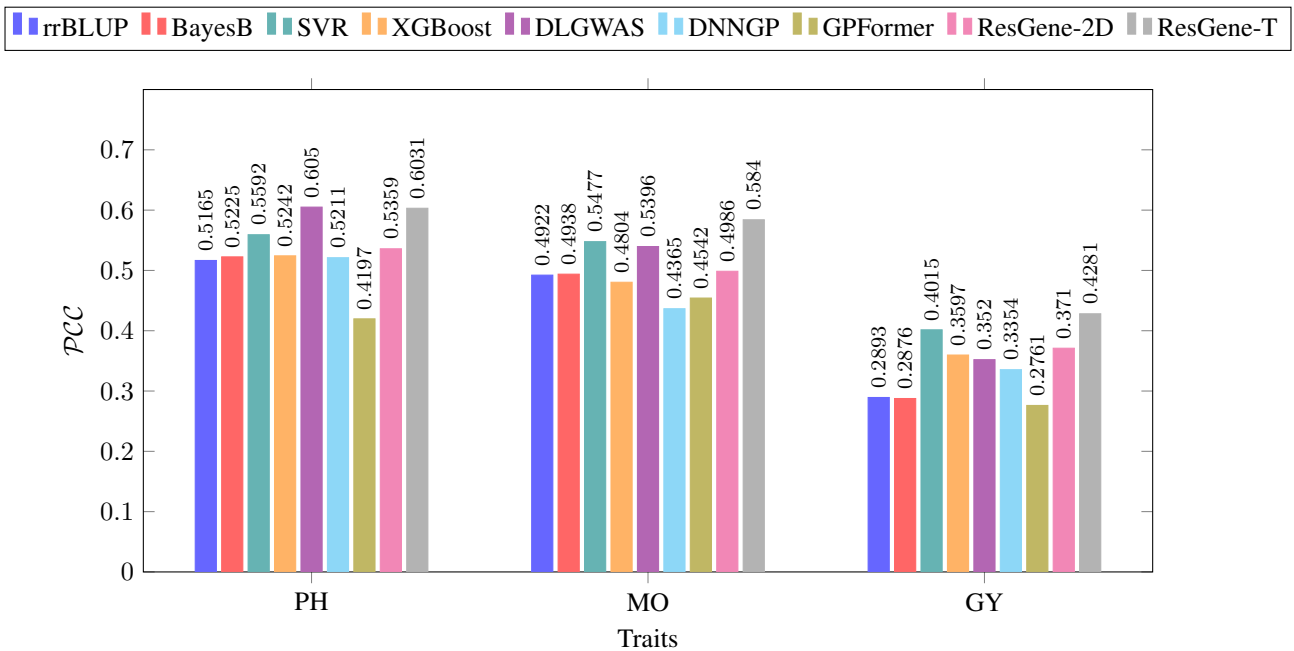

Fig.~\ref{fig:bar_icar} presents the performance of each model on the Soybean dataset. The X-axis shows the four phenotypic traits, and the Y-axis shows the corresponding $\mathcal{PCC}$ values for all nine models. Based on the average prediction accuracy across traits, ResGene-T achieves the highest accuracy $0.3252$. It outperforms rrBLUP, BayesB, SVR, XGBoost, DLGWAS, DNNGP, GPFormer, and ResGene-2D by $21.23\%$, $31.51\%$, $18.79\%$, $63.86\%$, $41.28\%$, $41.51\%$, $18.55\%$, and $3.55\%$ respectively.

For the Rice dataset, the results are presented in Fig.~\ref{fig:bar_rice}. Based on the average prediction accuracy, ResGene-T achieves the highest accuracy $0.4560$. It outperforms rrBLUP, BayesB, SVR, XGBoost, DLGWAS, DNNGP, GPFormer, and ResGene-2D by $59.58\%$, $34.65\%$, $20.52\%$, $1.74\%$, $6.74\%$, $68.25\%$, $57.30\%$, and $17.33\%$ respectively.

Fig.~\ref{fig:bar_sor} shows the results for the Sorghum dataset. On average, ResGene-T achieves the highest accuracy $0.5374$. It  outperforms rrBLUP, BayesB, SVR, XGBoost, DLGWAS, DNNGP, GPFormer, and ResGene-2D by $24.20\%$, $23.64\%$, $6.88\%$, $18.17\%$, $7.71\%$, $24.68\%$, $40.19\%$, and $14.70\%$ respectively.

\begin{table*}[ht]
\small
\centering
\caption{Comparative ranking of models across datasets and traits}
\label{tab:model_ranks}
\setlength{\tabcolsep}{4pt}
\renewcommand{\arraystretch}{1.3}
\begin{tabular}{|l|c|c|c|c|c|c|c|c|>{\columncolor{gray!15}}c|>{\columncolor{gray!15}}c|}
\hline
\textbf{Dataset}& \textbf{Trait} & \textbf{rrBLUP} & \textbf{BayesB} & \textbf{SVR} & \textbf{XGBoost} & \textbf{DLGWAS} & \textbf{DNNGP} & \textbf{GPFormer} & \textbf{ResGene-2D} & \textbf{ResGene-T} \\
\hline
\multirow{4}{*}{Soybean}& PH & 7 & 9 & 4 & 3 & 8 & 6 & 5 & 2 & 1  \\
& NN & 5 & 6 & 2 & 9 & 8 & 7 & 4 &  3& 1  \\
& GY & 1 & 2 & 6 & 7 & 7 & 8 & 5 & 4 & 3 \\
& CT & 7 & 9 & 5 & 8 & 4 & 6 & 3 &1 &2 \\
\hline
\multirow{3}{*}{Rice}& PH & 8 & 7 & 6 & 3 & 2 & 9 & 4 &5 &1 \\
& FT & 8 & 6 & 5 & 2 & 3 & 7 & 9 & 4 & 1 \\
& GY & 7 & 6 & 4 & 2 & 3 & 9 & 8 & 5& 1 \\
\hline
\multirow{3}{*}{Sorghum}& PH & 8 & 6 & 3 & 5 & 1 & 7& 9 &4 &2 \\
& MO & 6 & 5 & 2 & 7 & 3 & 9 & 8 & 4 &1 \\
& GY & 7 & 8 & 2 & 4 & 5 & 6 & 9 & 3 &1 \\

\noalign{\hrule height 1.2pt}
\multicolumn{2}{|c|}{Average $\mathcal{PCC}$}  & 0.3228 & 0.3309 & 0.3739 & 0.3503 & 0.3699 & 0.3025 &0.3117 & 0.3828 & 0.4281 \\
\hline
\multicolumn{2}{|c|}{Average Relative \% Gain}  & 32.61\% & 29.37\% & 14.51\% & 22.22\% & 15.73\% & 41.51\% & 37.35\% & 11.85\% &- \\
\hline
\multicolumn{2}{|c|}{Final Ranking}  & 1/10 & 0/10 & 0/10 & 0/10 & 1/10 & 0/10 &0/10 & 1/10 &7/10 \\
\hline
\multicolumn{2}{|c|}{Average Ranking}  & 6.4 & 6.4 & 3.9 & 5.2 & 4.4 & 7.4 & 6.4 & 3.5 &1.4\\
\hline
\end{tabular}

\end{table*}

We now examine the comparative performance of all nine models through statistical ranking analysis. Table~\ref{tab:model_ranks} summarizes the performance across multiple datasets and traits. For each trait within a dataset, the models are ranked based on their $\mathcal{PCC}$ values, where a lower rank indicates better predictive performance. In addition to per-trait rankings, the table reports aggregate statistics to facilitate comprehensive comparison. The row labeled Average $\mathcal{PCC}$ presents the mean $\mathcal{PCC}$ of each model computed across all traits. The Average Relative \% Gain shows the relative improvement of ResGene-T over competing methods based on average $\mathcal{PCC}$. The Final Ranking indicates how many times each model achieves the first position across all traits, expressed as a ratio. Finally, the Average Ranking is obtained by averaging the per-trait ranks of each model across all traits.

Table~\ref{tab:model_ranks} reveals that ResGene-T emerges as the best-performing model overall. Based on average $\mathcal{PCC}$, the models rank as follows: ResGene-T $(0.4281)$, ResGene-2D $(0.3828)$, SVR $(0.3739)$, DLGWAS $(0.3699)$, XGBoost $(0.3503)$, BayesB $(0.3309)$, rrBLUP $(0.3228)$, GPFormer $(0.3117)$, and DNNGP $(0.3025)$. This corresponds to relative improvements achieved by ResGene-T of $11.85\%$, $14.51\%$, $15.73\%$, $22.22\%$, $29.37\%$, $32.61\%$, $37.35\%$, and $41.51\%$ over ResGene-2D, SVR, DLGWAS, XGBoost, BayesB, rrBLUP, GPFormer, and DNNGP, respectively. 

ResGene-T achieves the highest accuracy in $7$ out of $10$ traits. Furthermore, ResGene-T achieves the lowest average rank of $1.4$ across all traits. In comparison comparatively, the second, third, fourth, and fifth best models are ResGene-2D, SVR, DLGWAS, XGBoost with the average ranks of $3.5$, $3.9$, $4.4$, and $5.2$ respectively. To demonstrate the rank of resGene is  statistically significantly different than other models, we apply the Friedman rank test on this. The test yields a $p$-value of $0.000008$, revealing that the rank of ResGene-T is clearly significantly higher. 

\section{Conclusion}\label{sec:con}

 Traditionally, the use of DL in GP passes the genotype as a sequence of characters to a 1D-CNN. Based on a similar work of classification problems, we propose ResGene-2D which converts the genotype into a 2D image and feeds it to a 2D-CNN. The idea here is that this transformation captures biological interactions between SNPs more effectively. This approach does capture these interactions well, however, it takes all the layers of the CNN to do so (since the whole image is read in the last layer). Thus, the model is not trained well. Hence, we propose a novel idea of converting the 2D image into a 3D/tensor representation and feeding it to a 2D-CNN. This ensures that the whole genotype is read in the earlier layers of the model, leading to better learning of biological interactions among the SNPs and subsequently giving better results.

We evaluate ResGene-2D and ResGene-T on three major crop species, namely soybean, rice, and sorghum, across ten phenotypic traits. We compare both our models against seven popular methods spanning statistical, machine learning, and deep learning models. Between our two proposed models, ResGene-T outperforms ResGene-2D by an average $\mathcal{PCC}$ improvement of $11.85\%$. When compared against the other seven methods, ResGene-T achieves average $\mathcal{PCC}$ improvements ranging from $14.51\%$ to $41.51\%$. ResGene-T delivers the best performance on $7$ out of $10$ traits with an average rank of $1.4$. In comparison, the next two best methods are ResGene-2D and SVR with average ranks of $3.5$, and
$3.9$, respectively. These results confirm the effectiveness and consistency of our proposed models.

While ResGene-2D and ResGene-T establish a strong foundation for GP, several limitations naturally point toward promising future directions. First, a deeper theoretical analysis of the proposed framework could provide fundamental insights into its behavior, where reformulating GP within the framework of linear solvers may establish connections to Krylov subspace methods for improved computational scalability~\cite{choudhary2018stability}. Second, extending ResGene-2D and ResGene-T for multi-trait prediction represents a promising direction for improving predictive reliability~\cite{ahuja2022multigoal}. Third, incorporating implicit modeling strategies to better capture the relationship between genotypic and phenotypic data remains an important avenue for future work~\cite{kim2005effectiveness}. Fourth, exploring approximate computing paradigms could further enhance computational efficiency in large scale GP tasks~\cite{ullah2020l2l}. By acknowledging these limitations as natural extensions, we emphasize that ResGene provides a solid foundation for future advancements in GP.

\section*{Acknowledgments}
The authors sincerely thank Mr. Aayush Dhanesh Agrawal, who was an M.Tech. student on this project for the initial experimentation. The authors also are very thankful Dr. Milind B. Ratnaparkhe, Principal Scientist at ICAR-NSRI Indore, for providing the Soybean dataset.

\bibliographystyle{ieeetr}
\bibliography{ref}

@article{azodi2019benchmarking,
  title={{Benchmarking Parametric and Machine Learning Models for Genomic Prediction of Complex Traits}},
  author={Azodi, Christina B and Bolger, Emily and McCarren, Andrew and Roantree, Mark and de Los Campos, Gustavo and Shiu, Shin-Han},
  journal={G3: Genes, Genomes, Genetics},
  volume={9},
  number={11},
  pages={3691--3702},
  year={2019},
  publisher={Oxford University Press}
}

@inproceedings{lu2025soybean,
  title={{Soybean genomic phenotype prediction method based on improving the transformer model with batch normalization and cosine annealing algorithm}},
  author={Lu, Xiaolian and Liu, Changhua and Wang, Jing},
  booktitle={International Conference on Artificial Intelligence and Machine Learning Research (CAIMLR 2024)},
  volume={13635},
  pages={227--233},
  year={2025},
  organization={SPIE}
}

@inproceedings{pandey2023resnet,
  title={{ResNet-18 comparative analysis of various activation functions for image classification}},
  author={Pandey, Gaurav Kumar and Srivastava, Sumit},
  booktitle={2023 International Conference on Inventive Computation Technologies (ICICT)},
  pages={595--601},
  year={2023},
  organization={IEEE}
}

@article{liu2019phenotype,
  title={{Phenotype Prediction and Genome-Wide Association Study Using Deep Convolutional Neural Network of Soybean}},
  author={Liu, Yang and Wang, Duolin and He, Fei and Wang, Juexin and Joshi, Trupti and Xu, Dong},
  journal={Frontiers in Genetics},
  volume={10},
  pages={1091},
  year={2019},
  publisher={Frontiers Media SA}
}

@article{zhang2010best,
  title={{Best Linear Unbiased Prediction of Genomic Breeding Values Using a Trait-Specific Marker-Derived Relationship Matrix}},
  author={Zhang, Zhe and Liu, Jianfeng and Ding, Xiangdong and Bijma, Piter and de Koning, Dirk-Jan and Zhang, Qin},
  journal={PloS One},
  volume={5},
  number={9},
  pages={e12648},
  year={2010},
  publisher={Public Library of Science San Francisco, USA}
}

@misc{icarNSRI,
  title        = {{ICAR-National Soybean Research Institute, Indore}},
  howpublished = {\url{https://icar-nsri.res.in/}},
  note         = {Accessed: 10 November 2025},
  year         = {2025},
  key          = {ICAR-NSRI}
}

@article{yan2021lightgbm,
  title={{LightGBM: accelerated genomically designed crop breeding through ensemble learning}},
  author={Yan, Jun and Xu, Yuetong and Cheng, Qian and Jiang, Shuqin and Wang, Qian and Xiao, Yingjie and Ma, Chuang and Yan, Jianbing and Wang, Xiangfeng},
  journal={Genome Biology},
  volume={22},
  number={1},
  pages={271},
  year={2021},
  publisher={Springer}
}

@article{gaptiv3,
    author = {Wang, Jiabo and Zhang, Zhiwu},
    title = {{GAPIT Version 3: Boosting Power and Accuracy for Genomic Association and Prediction}},
    journal = {Genomics, Proteomics \& Bioinformatics},
    volume = {19},
    number = {4},
    pages = {629-640},
    year = {2021},
    month = {09},
    issn = {1672-0229},
    doi = {10.1016/j.gpb.2021.08.005},
    url = {https://doi.org/10.1016/j.gpb.2021.08.005},
    eprint = {https://academic.oup.com/gpb/article-pdf/19/4/629/57581860/gpb_19_4_629.pdf},
}

@article{muneeb2022can,
  title={{Can we convert genotype sequences into images for cases/controls classification?}},
  author={Muneeb, Muhammad and Feng, Samuel F and Henschel, Andreas},
  journal={Frontiers in Bioinformatics},
  volume={2},
  pages={914435},
  year={2022},
  publisher={Frontiers Media SA}
}

@article{wang2025wheatgp,
  title={{WheatGP, a genomic prediction method based on CNN and LSTM}},
  author={Wang, Chunying and Zhang, Di and Ma, Yuexin and Zhao, Yonghao and Liu, Ping and Li, Xiang},
  journal={Briefings in Bioinformatics},
  volume={26},
  number={2},
  pages={bbaf191},
  year={2025},
  publisher={Oxford University Press}
}

@article{wang2023dnngp,
  title={{DNNGP, a deep neural network-based method for genomic prediction using multi-omics data in plants}},
  author={Wang, Kelin and Abid, Muhammad Ali and Rasheed, Awais and Crossa, Jose and Hearne, Sarah and Li, Huihui},
  journal={Molecular Plant},
  volume={16},
  number={1},
  pages={279--293},
  year={2023},
  publisher={Elsevier}
}

@article{islam2023deepcgp,
  title={{DeepCGP: A Deep Learning Method to Compress Genome-Wide Polymorphisms for Predicting Phenotype of Rice}},
  author={Islam, Tanzila and Kim, Chyon Hae and Iwata, Hiroyoshi and Shimono, Hiroyuki and Kimura, Akio},
  journal={IEEE/ACM Transactions on Computational Biology and Bioinformatics},
  volume={20},
  number={3},
  pages={2078--2088},
  year={2023},
  publisher={IEEE}
}

@article{wu2023transformer,
  title={{A transformer-based genomic prediction method fused with knowledge-guided module}},
  author={Wu, Cuiling and Zhang, Yiyi and Ying, Zhiwen and Li, Ling and Wang, Jun and Yu, Hui and Zhang, Mengchen and Feng, Xianzhong and Wei, Xinghua and Xu, Xiaogang},
  journal={Briefings in Bioinformatics},
  volume={25},
  number={1},
  year={2023},
  publisher={Oxford Academic}
}

@article{mao2025mdnn,
  title={{MDNN: memetic deep neural network for genomic prediction}},
  author={Mao, Yijun and Peng, Xingcheng and Weng, Jian and Jiang, Rongjin and Kuang, Yingjie and Weng, Jiasi and Pang, Rui and Xiong, Yunyan and Gu, Wanrong and Tang, Deyu},
  journal={Briefings in Bioinformatics},
  volume={26},
  number={4},
  pages={bbaf352},
  year={2025},
  publisher={Oxford University Press}
}

@article{haile2021genomic,
  title={{Genomic prediction of agronomic traits in wheat using different models and cross-validation designs}},
  author={Haile, Teketel A and Walkowiak, Sean and N’Diaye, Amidou and Clarke, John M and Hucl, Pierre J and Cuthbert, Richard D and Knox, Ron E and Pozniak, Curtis J},
  journal={Theoretical and Applied Genetics},
  volume={134},
  number={1},
  pages={381--398},
  year={2021},
  publisher={Springer}
}

@article{meher2022performance,
  title={{Performance of Bayesian and BLUP alphabets for genomic prediction: analysis, comparison and results}},
  author={Meher, Prabina Kumar and Rustgi, Sachin and Kumar, Anuj},
  journal={Heredity},
  volume={128},
  number={6},
  pages={519--530},
  year={2022},
  publisher={Springer International Publishing Cham}
}

@article{grinberg2020evaluation,
  title={{An evaluation of machine-learning for predicting phenotype: studies in yeast, rice, and wheat}},
  author={Grinberg, Nastasiya F and Orhobor, Oghenejokpeme I and King, Ross D},
  journal={Machine Learning},
  volume={109},
  number={2},
  pages={251--277},
  year={2020},
  publisher={Springer}
}

@article{sehrawat2023predicting,
  title={{Predicting phenotypes from novel genomic markers using deep learning}},
  author={Sehrawat, Shivani and Najafian, Keyhan and Jin, Lingling},
  journal={Bioinformatics Advances},
  volume={3},
  number={1},
  pages={vbad028},
  year={2023},
  publisher={Oxford University Press}
}

@article{ma2018deep,
  title={{A deep convolutional neural network approach for predicting phenotypes from genotypes}},
  author={Ma, Wenlong and Qiu, Zhixu and Song, Jie and Li, Jiajia and Cheng, Qian and Zhai, Jingjing and Ma, Chuang},
  journal={Planta},
  volume={248},
  number={5},
  pages={1307--1318},
  year={2018},
  publisher={Springer}
}

@article{lorenz2011genomic,
  title={{Genomic Selection in Plant Breeding: Knowledge and Prospects}},
  author={Lorenz, Aaron J and Chao, Shiaoman and Asoro, Franco G and Heffner, Elliot L and Hayashi, Takeshi and Iwata, Hiroyoshi and Smith, Kevin P and Sorrells, Mark E and Jannink, Jean-Luc},
  journal={Advances in Agronomy},
  volume={110},
  pages={77--123},
  year={2011},
  publisher={Elsevier}
}

@article{ghosh2021enriched,
  title={{Enriched Random Forest for High Dimensional Genomic Data}},
  author={Ghosh, Debopriya and Cabrera, Javier},
  journal={IEEE/ACM Transactions on Computational Biology and Bioinformatics},
  volume={19},
  number={5},
  pages={2817--2828},
  year={2021},
  publisher={IEEE}
}

@article{demvsar2006statistical,
  title={{Statistical Comparisons of Classifiers over Multiple Data Sets}},
  author={Dem{\v{s}}ar, Janez},
  journal={Journal of Machine Learning Research},
  volume={7},
  number={Jan},
  pages={1--30},
  year={2006}
}

@article{andrade2019p,
  title={{The P Value and Statistical Significance: Misunderstandings, Explanations, Challenges, and Alternatives}},
  author={Andrade, Chittaranjan},
  journal={Indian Journal of Psychological Medicine},
  volume={41},
  number={3},
  pages={210--215},
  year={2019},
  publisher={SAGE Publications Sage India: New Delhi, India}
}

@article{feng2023survey,
  title={{A survey of visual neural networks: current trends, challenges and opportunities}},
  author={Feng, Ping and Tang, Zhenjun},
  journal={Multimedia Systems},
  volume={29},
  number={2},
  pages={693--724},
  year={2023},
  publisher={Springer}
}

@article{wang2022using,
  title={{Using machine learning to improve the accuracy of genomic prediction of reproduction traits in pigs}},
  author={Wang, Xue and Shi, Shaolei and Wang, Guijiang and Luo, Wenxue and Wei, Xia and Qiu, Ao and Luo, Fei and Ding, Xiangdong},
  journal={Journal of Animal Science and Biotechnology},
  volume={13},
  number={1},
  pages={60},
  year={2022},
  publisher={Springer}
}

@article{wang2018expanding,
  title={{Expanding the BLUP alphabet for genomic prediction adaptable to the genetic architectures of complex traits}},
  author={Wang, Jiabo and Zhou, Zhengkui and Zhang, Zhe and Li, Hui and Liu, Di and Zhang, Qin and Bradbury, Peter J and Buckler, Edward S and Zhang, Zhiwu},
  journal={Heredity},
  volume={121},
  number={6},
  pages={648--662},
  year={2018},
  publisher={Springer International Publishing Cham}
}

@inproceedings{kim2005effectiveness,
  title={Effectiveness of Implicit Rating Data on Characterizing Users in Complex Information Systems},
  author={Kim, Seonho and Murthy, Uma and Ahuja, Kapil and Vasile, Sandi and Fox, Edward A},
  booktitle={International Conference on Theory and Practice of Digital Libraries},
  pages={186--194},
  year={2005},
  organization={Springer}
}

@article{choudhary2018stability,
  title={{Stability analysis of Bilinear Iterative Rational Krylov Algorithm}},
  author={Choudhary, Rajendra and Ahuja, Kapil},
  journal={Linear Algebra and its Applications},
  volume={538},
  pages={56--88},
  year={2018},
  publisher={Elsevier}
}

@article{ahuja2022multigoal,
  title={{Multigoal-oriented error estimation and mesh adaptivity for fluid--structure interaction}},
  author={Ahuja, Kapil and Endtmayer, Bernhard and Steinbach, Marc Christian and Wick, Thomas},
  journal={Journal of Computational and Applied Mathematics},
  volume={412},
  pages={114315},
  year={2022},
  publisher={Elsevier}
}

@inproceedings{ullah2020l2l,
  title={{L2L: A Highly Accurate Log\_2\_Lead Quantization of Pre-trained Neural Networks.}},
  author={Ullah, Salim and Gupta, Siddharth and Ahuja, Kapil and Tiwari, Aruna and Kumar, Akash},
  booktitle={DATE},
  pages={979--982},
  year={2020}
}
\vspace{-7cm}
\begin{IEEEbiography}[{\includegraphics[width=1in,height=1.25in,clip,keepaspectratio]{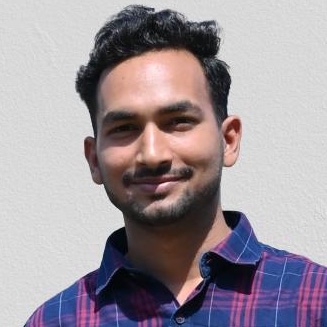}}]
{Kuldeep Pathak} received his Bachelor’s degree from Acharya Narendra Deva University of Agriculture and Technology (ANDUAT), India. After that, he worked as a Systems Engineer at Tata Consultancy Services (TCS) before starting his Ph.D. in 2023. He is working closely with Leibniz University Hannover (Germany), where he has spent two research visits during the winter terms. His research interests include machine learning and artificial intelligence and their applications in plant domain.

\end{IEEEbiography}
\vspace{-7cm}
\begin{IEEEbiography}[{\includegraphics[width=1in,height=1.25in,clip,keepaspectratio]{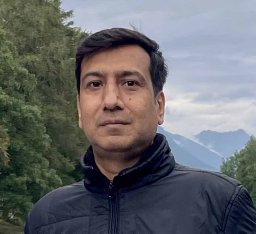}}]
{Kapil Ahuja}
after completing a double Master’s degree and a Ph.D. from Virginia Tech (USA), 
Prof. Ahuja pursued a postdoctoral fellowship at the Max Planck Institute, 
Magdeburg, Germany. Subsequently, he served as an Assistant Professor and 
Associate Professor in the Department of Computer Science and Engineering 
at IIT Indore, where he is currently a Full Professor.
He has also held visiting positions at Brown University (USA), IISc (India), 
UT Austin (USA), IMT Atlantique (France), Sandia National Laboratories (USA), 
TU Dresden (Germany), and TU Braunschweig (Germany). His research interests include applied machine learning and numerical linear algebra.

\end{IEEEbiography}
\vspace{-7cm}
\begin{IEEEbiography}[{\includegraphics[width=1in,height=1.25in,clip,keepaspectratio]{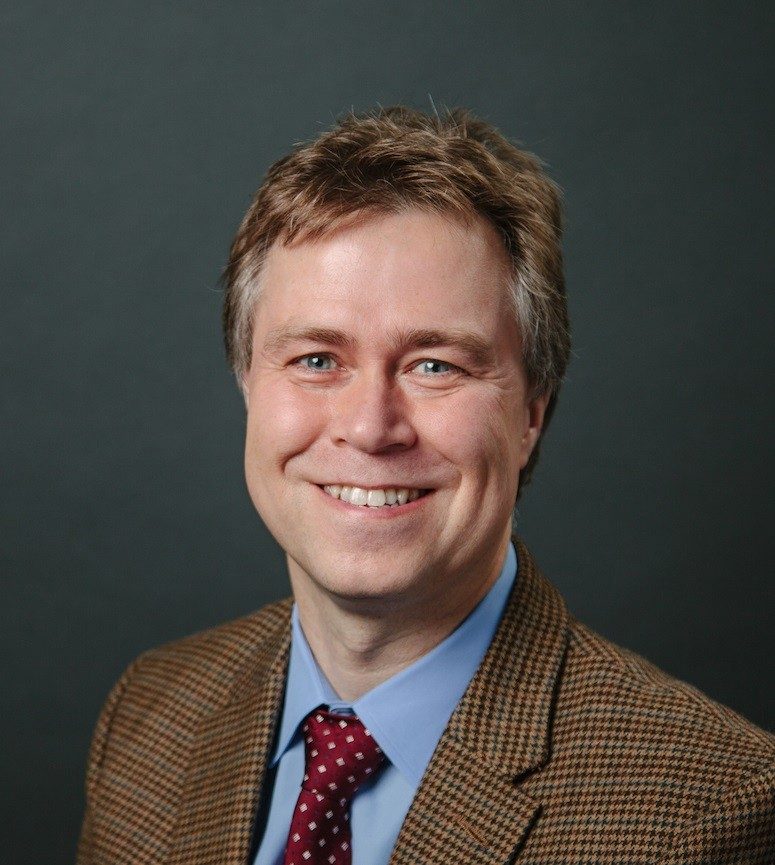}}]
{Eric de Sturler} received his Ph.D. degree from the Technische Universiteit Delft (Netherlands). Following his doctoral studies, he held research positions at the ETHZ in Switzerland. He subsequently served on the faculty of the department of Computer Science at the University of Illinois at Urbana-Champaign. He is currently a Professor in the Department of Mathematics at Virginia Tech. He also serves as the Director of the Academy of Data Science at Virginia Tech. His research interests include numerical analysis and data science.

\end{IEEEbiography}

\end{document}